\documentclass[conference]{IEEEtran}
\IEEEoverridecommandlockouts

\usepackage{multirow}
\usepackage{cite}
\usepackage{url}
\usepackage{footnote}
\usepackage{amsmath,amssymb,amsfonts}
\usepackage{algorithmic}
\usepackage{graphicx}
\usepackage{textcomp}
\usepackage{xcolor}
\def\BibTeX{{\rm B\kern-.05em{\sc i\kern-.025em b}\kern-.08em
    T\kern-.1667em\lower.7ex\hbox{E}\kern-.125emX}}
\begin{document}

\title{Gaining a Sense of Touch. Physical Parameters Estimation using a Soft Gripper and Neural Networks}

\author{\IEEEauthorblockN{1\textsuperscript{st} Micha{\l} Bednarek}
\IEEEauthorblockA{\textit{Institute of Robotics and Machine Intelligence} \\
\textit{Poznan University of Technology}\\
Poznan, Poland\\michal.bednarek@put.poznan.pl}
\and
\IEEEauthorblockN{2\textsuperscript{nd} Piotr Kicki}
\IEEEauthorblockA{\textit{Institute of Robotics and Machine Intelligence} \\
\textit{Poznan University of Technology}\\
Poznan, Poland}
\and
\IEEEauthorblockN{3\textsuperscript{rd} Jakub Bednarek}
\IEEEauthorblockA{\textit{Institute of Robotics and Machine Intelligence} \\
\textit{Poznan University of Technology}\\
Poznan, Poland}
\and
\IEEEauthorblockN{4\textsuperscript{th} Krzysztof Walas}
\IEEEauthorblockA{\textit{Institute of Robotics and Machine Intelligence} \\
\textit{Poznan University of Technology}\\
Poznan, Poland}
}

\maketitle

\begin{abstract}
Soft grippers are gaining significant attention in the manipulation of elastic objects, where it is required to handle soft and unstructured objects which are vulnerable to deformations. A crucial problem is to estimate the physical parameters of a squeezed object to adjust the manipulation procedure, which is considered as a significant challenge. To the best of the authors' knowledge, there is not enough research on physical parameters estimation using deep learning algorithms on measurements from direct interaction with objects using robotic grippers. In our work, we proposed a trainable system for the regression of a stiffness coefficient and provided extensive experiments using the physics simulator environment. Moreover, we prepared the application that works in the real-world scenario. Our system can reliably estimate the stiffness of an object using the Yale OpenHand soft gripper based on readings from Inertial Measurement Units (IMUs) attached to its fingers. Additionally, during the experiments, we prepared three datasets of signals gathered while squeezing objects -- two created in the simulation environment and one composed of real data. 
\end{abstract}


\section{Introduction}  

\begin{figure}[htbp]
\centerline{\includegraphics[width=0.45\textwidth]{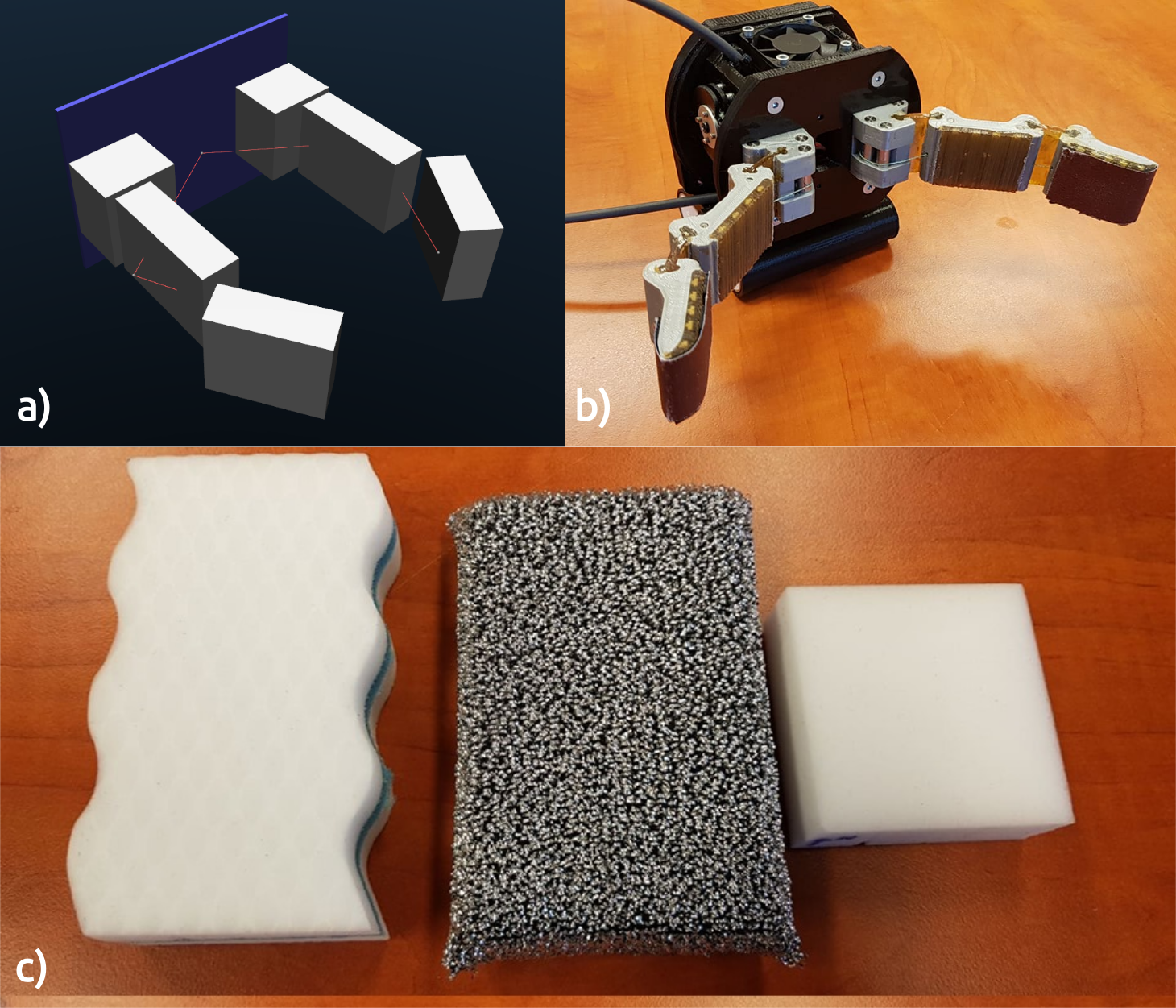}}
\caption{To test our system, we prepared a real-world scenario using a 2-finger Yale OpenHand gripper~\cite{YaleOpenHand}. To provide a sufficient number of training samples for the learning process, we modeled the gripper in the MuJoCo simulator as it is depicted in \textit{a)}. In \textit{b)} real fingers consist of three plastic blocks with flexible parts made of urethane.  In~\textit{c)}, there are presented examples of sponges, exposing different stiffness, used in our real-world experiments.}
\label{fig:real_experiment}
\end{figure}

Humans have an innate ability to perceive the physics of the world around them. As we are biologically equipped with a very sophisticated sensory system that delivers data to the brain, no-one deliberately plans how to grab a cup of tea, squeeze a wet sponge or flip a book page. We all \textit{know} how to do that and how to predict deformations of different objects based on their physical properties. Moreover, humans have at their disposal soft and highly effective grippers -- hands. Taking into account our assumptions about the world that come from our brains combined with the \textit{embodied intelligence}~\cite{IIda07} of our hands, we can flawlessly adjust the process of manipulation to fluctuating external conditions. However, machines do not have such an in-built proficiency. Thus, their ability to manipulate only allows for handling repetitive tasks and forbids them to adapt to new types of objects efficiently.

Biologically inspired soft grippers~\cite{YaleOpenHand, MIT3fingerGripper, MITorigamiGripper, LaschiGripper} are designed to handle not only rigid bodies but also deformable and frequently delicate objects. How they can interact with the real-world and how they can adjust to different objects is ruled by their property called \textit{intelligence by mechanics}~\cite{IIda07}. One can observe a significant rise in the number of available applications of sensors capable of capturing high-dimensional deformations of soft and unpredictable systems ~\cite{BatchFabricationSensor, TactileTwo, Realumuto}. However, in our work, we state that traditional and widespread solid-state sensors can also be successfully used to predict the physical nature of the robot's surroundings. Thereby, we propose a hybrid approach that connects an \textit{embodied intelligence} of a soft gripper with an \textit{artificial intelligence} system to provide an easy to use, open-source and inexpensive method of estimating the physical properties of objects with various stiffness parameters.

The following study presents the deep learning, real-world application for stiffness coefficient estimation based on data from Inertial Measurement Units (IMUs) attached to the fingers of the gripper. Our contribution are:

\begin{enumerate}
    \item Creation of simulated environments for generating data and examining the soft gripper in various scenarios.
    \item Verification of the performance of three architectures of neural networks in the task of stiffness parameter estimation -- purely convolutional and two recurrent models.
    \item The real-world application with a thorough examination of closing the reality-gap in the machine learning system.
    \item The open-source implementation and data used in the experiments available online~\footnote{\url{https://github.com/mbed92/soft-grip}}.
\end{enumerate}

To prepare the real-world experiment, we used a two-finger gripper based on the Yale OpenHand Project~\cite{YaleOpenHand} with two IMUs attached to its fingers. The motivation standing behind the choice of that type of sensor is twofold. First of all, typically soft grippers have no hinges and do not use encoders, therefore we cannot track their movement directly. Following the research on the Pisa/IIT SoftHand~\cite{pisa_hand_imu_justification}, the IMU signal is suitable while tracking the movement of under-actuated and elastic fingers of the gripper. Secondly, IMUs are inexpensive, small and widespread among the robotics community. The course of the research is as follows: first, experiments were carried out exclusively on data from the physics simulator to verify the capabilities of both architectures and examine the generalization of the stiffness parameter regression between different shapes of squeezed objects. Thereafter, we investigated the well-known machine learning problem -- closing the reality gap between the simulation and real-world data. In our experiments, we exploited the MuJoCo~\cite{mujoco} simulator to provide a sufficient number of training samples. The IMU device model used in our work is the MPU-9250 model. In the Figure~\ref{fig:real_experiment} there was presented the setup used in the real-world scenario with its simulation model and exemplary objects.

The remainder of the paper organized as follows. First, we will review the state-of-the-art in the field of physical parameters estimation from haptic data. Then, we will provide a description of our experiments and prepared setups. Next, we will move on to the results section followed by the discussion. Finally, concluding remarks will be given.

\section{Related work}  

In this section, we provided a comprehensive literature review both on the approaches to measuring and estimating the object stiffness. Further, we showed current advances in processing data from IMUs for a wide range of purposes.

\subsection{Stiffness Measurements and Estimation Techniques}
Knowledge about material's stiffness is highly demanded in many practical applications such as industrial robotics, where a robot may use this information to predict an object's deformation. We presented current advances in finding object stiffness in two general approaches: \textit{measurement}, where the result was obtained with the usage of advanced, dedicated sensors and \textit{estimation}, where we focused on the possible use of all available information relevant for a given task.

\textbf{Measurement} --
In~\cite{rails} authors proposed a method for continuous rail stiffness measurements using the accelerometer and oscillating mass on the rolling wheel. Non-contact measurement of spindle stiffness was presented in~\cite{noncontact_stiff}. The authors proposed a magnetic loading device that enables one to perform the measurement while spindle rotates. Due to the usage of magnetic loading, that method is limited to the ferromagnetic objects. Measuring the stiffness is also possible at a much smaller scale. The authors~of~\cite{nano} presented the review of the nanoindentation continuous stiffness measurement technique and its applications. The range of stiffness coefficients of materials is extensive. To avoid saturation and enhance precision authors of~\cite{stiff_measurement_system} proposed a portable measurement tool able to adjust the sensing range by manipulating tool parameters, such as touch module separation, indenter protrusion and spring constant of the force sensing module. Authors of~\cite{foams, foams2} analyzed the stiffness measurement techniques applied to the polymer foams, which are cognate to those used in this paper. In~\cite{foams} a procedure for measuring the stiffness of the object using dot markers on the object and compression plates to exert the force on the object was proposed. Authors stress the fact, that non-axial compression tests result in worse performance, but it is usually the case in robotic manipulation. In~\cite{pisa_hand_imu_justification} authors proposed the IMU-based approach to reconstruct the configuration of soft gripper.

\textbf{Estimation} --
A method that does not require measuring the object deformation was proposed in~\cite{stiff_from_force}. The authors proposed Candidate Observer-Based Algorithm, which exploits two force observers, with different stiffness candidates, for estimating the stiffness of objects with complicated geometry. Unfortunately, authors did not refer their method to the ground truth stiffness measurements. However, such a comparison was made in~\cite{NN_stiff}, where the neural network was trained to predict the stiffness coefficient based on the maximum penetration and the maximum contact pressure variation. An alternative deep learning approach for understanding the haptic properties of objects was proposed in~\cite{haptic_from_vision_and_biotac}. The real-world objects were classified in the set of haptic adjectives in the multi-label fashion based on haptic signals from BioTac sensors~\cite{biotac_sensor} and images. The method for object stiffness estimation was proposed in~\cite{SPA_xD}. Authors used small optical force sensors mounted on the fingertips, a known kinematic model of the robotic hand and a vision system to calculate the stiffness based on the force and displacement readings.

\subsection{IMU Measurements Applications}

A significant advantage of the IMU sensor is its easy availability and low price. These features resulted in the popularity in many robotics applications. One of them is a robot's state estimation. In~\cite{humanoid}, acceleration and angular velocities collected from sensors located on the humanoid leg, together with joints positions were used to estimate the velocity of links. Authors in~\cite{GRF} presented multiple interesting approaches to measure indirectly the ground reaction forces during the human walk with the use of wearable IMUs. The other field where the acceleration can be utilized is a material classification. In~\cite{penn_haptix} authors used the haptic device SensAble Phantom Omni~\cite{haptic_device} to gather the accelerations and velocities while scratching the material surfaces. That dataset was used in~\cite{scratching}, where a deep convolutional neural network was learned to map from raw signals to classes of textures. The presented method stays close to our solution. However, in our work, we performed regression instead of classification.

\section{Method}
\label{sec:matandmet}  

In the following section, we described the experimental design and detailed information about both real-world and simulated environments for our experiments. Then, we proposed architectures for the comparison of deep learning models used in our research.

\subsection{Experimental Design}
\label{sec:sec:experiment_design}

The performance of Neural Networks (NN) was verified using a k-fold cross-validation technique in each experiment. That method assesses the error rate and generalization ability of predictive models. In our research, it proceeds as follows: shuffle the dataset, then split the dataset into k subsets (folds), proceed with training using the k - 1 folds of data and validate the performance at the end of an epoch using the k-th fold. Additionally, unless otherwise stated, after each epoch we test the current NN model using separate test data. After that, the procedure is repeated by starting the training of a neural network from scratch on other folds of data. In our research, to ensure a fair comparison of trained NNs, we did the 5-fold cross-validation for all experiments. As we perform the regression task, we chose a Mean Absolute Error (MAE) and a Mean Absolute Percentage Error (MAPE) as performance metrics, to verify both absolute and relative errors. Considering the usage of the cross-validation technique, in the following description of datasets we provided the number of samples in the training-validation sets together, and separately for test sets if needed. The summary of all datasets used in our experiments was presented in Table~\ref{tab:dataset_all}.

\begin{table}[hbt]
\centering
\caption{The number of samples in datasets used in our experiments based on the cross-validation.}
\begin{tabular}{|c|c|c|}
\hline
\textbf{Name} & \textbf{Train / Validation} & \textbf{Test} \\ \hline
\textbf{Simulation (box only)} & 5000 & - \\ \hline
\textbf{Simulation (all shapes)} & 3999 & 133 $\times$ 3 \\ \hline
\textbf{Real-world} & 200 & 300 \\ \hline
\end{tabular}
\label{tab:dataset_all}
\end{table}

\subsection{Real Data}
\label{sec:sec:real_data}

The Yale OpenHand shown in Figure~\ref{fig:real_experiment} is the under-actuated, two-finger soft gripper with joints in the form of urethane elements to assure the elasticity of fingers. The real-world model was 3D printed and driven by hobby servos capable of generating a force up to 10N. The hand had IMU mounted at the fingertips of the hand. The IMU signals were used to estimate stiffness. In our work, we assessed how the \textit{embodied intelligence} of such soft gripper could be used alongside with the \textit{artificial intelligence} system to predict the real stiffness coefficient of a squeezed object. In the following section the real-world data gathering process was presented.

First we estimated the stiffness coefficient for real objects in the dataset. To calculate ground-truth values of the stiffness coefficient of real-world objects, we used the Universal Robot UR3 collaborative manipulator, which was able to measure torques and forces in its joints and tool respectively. The robot had 3d printed plastic bar mounted at the flange. Using the Dynamic Force Control mode and pressing objects with the desired force we were able to accurately measure the displacement under specific force from robot state readings. Thus, the stiffness parameter was computed according to the Equation~\ref{eq:stiffness}, where $F_1$ and $F_2$ are forces in Z-axis while pressing an object with a tool and $\lvert d_1 - d_2\rvert$ is the relative distance that correspond to the deformations under $F_1$ and $F_2$. In our work, we assume that the estimated stiffness parameter is homogeneous for the entire object. Table~\ref{tab:objects} contains stiffness coefficients measured experimentally for each object.

\begin{equation}
    \label{eq:stiffness}
    k = \frac{\lvert F_1 - F_2 \rvert}{\lvert d_1 - d_2\rvert}
\end{equation}

\begin{table}[htbp]
\centering
\caption{Stiffness coefficients computed for 5 different real objects.}
\begin{tabular}{|c|c|}
\hline
\textbf{Object} & \textbf{Stiffness {[}N/m{]}} \\ \hline
Wire sponge     & 909                  \\ \hline
Hard sponge     & 1020                  \\ \hline
Polish sponge   & 735                  \\ \hline
Soft sponge     & 380                  \\ \hline
Squash ball     & 1353                  \\ \hline
\end{tabular}
\label{tab:objects}
\end{table}

After estimating the value of the ground-truth stiffness coefficients, we used Yale OpenHand and collected signals for each object from the squeezing motion. In total, we gathered 500 series. They consist of 12 sensor readings (2 $\times$ IMU readings: [$Acc_{x}$, $Acc_{y}$, $Acc_{z}$, $\omega_{x}$, $\omega_{y}$, $\omega_{z}$]) each 200 time steps long. All samples are equally distributed among the objects -- 100 samples per each object. The data was split into two subsets -- 200 train and 300 test samples that were used in sim-to-real experiments. Both sets in all our experiments remain unchanged, thus test data is never used in the NN training.

However, the diversity of parameters in the real dataset was minimal. In that situation, the task of regression could inevitably and implicitly turn into a classification. To overcome that problem, we prepared a second dataset based on the simulation. The parameters of a stiffness coefficient were adjusted to meet measured values.

\subsection{Simulation}

\begin{figure}[htbp]
\centerline{\includegraphics[width=0.45\textwidth]{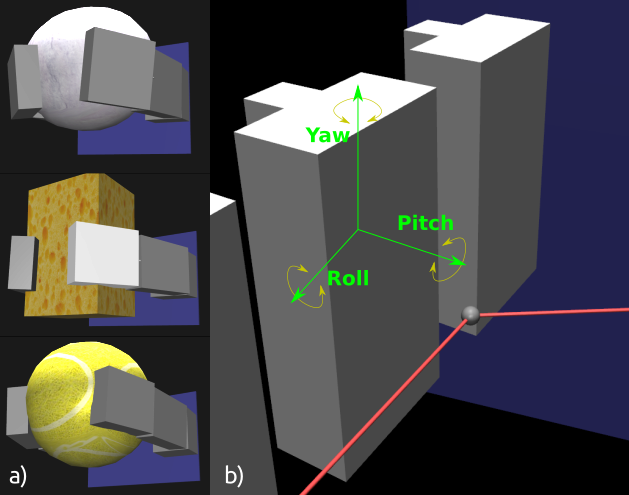}}
\caption{Soft-robotic gripper in the MuJoCo environment: \textit{a)} the gripper squeezes and releases objects in three shapes - a ball, box and a cylinder, all with a changing stiffness parameter; \textit{b)} each geometrical block of each finger is connected to others by three hinges. In this setup, we can easily adjust the ranges of each joint in a roll, pitch and yaw axes.}
\label{fig:scene_gripper_sim}
\end{figure}

Modern neural networks frequently suffer from the limited ability to generalize to new domains which are out of their training dataset. However, the rising popularity of machine learning techniques in the robotics community leads to a significantly increased need for data from a variety of experiments. To fulfill that demand, the state of the art approach is to perform experiments in simulation and use them to feed neural networks. In the case of tasks which involve physical interaction, researchers can choose from a wide range of available physics simulators. In our case, we selected MuJoCo physics simulator, due to its new features regarding soft objects modeling. The simulated soft-robotic gripper was shown in Figure~\ref{fig:scene_gripper_sim}. Fingers were connected by tendons and they are pulled by the actuator, which simulates the pneumatic cylinder. Our model was based on the 3-finger real gripper~\cite{MIT3fingerGripper}. As it was depicted in Figure~\ref{fig:scene_gripper_sim}a, during experiments, our gripper squeezed and released objects of three shapes - a ball, box and a cylinder, all with a changing stiffness parameter. To simulate elastic deformations of the gripper, each geometrical block of each finger is connected to others by three hinges. In this setup, we can easily adjust the ranges of each joint in a roll, pitch and yaw axes, as was depicted in the Figure~\ref{fig:scene_gripper_sim}b. Finally, each 8-block finger behaves similarly to the elastic finger.

A stiffness coefficient in all our experiments (including the real-world scenario) is defined in the same way as in the MuJoCo simulator -- the \textit{softness} of an object is characterized as the stiffness of springs attached from one side to the geometrical blocks on a surface and from the other side to the center of it. We always assume that the object is homogeneous.

The process of data collection is designed as follows. An object is located between fingers and the actuator starts to close the gripper to squeeze the object. After a half of an episode, the gripper opens. During the process, an object is embraced by fingers that adapt themselves to its shape. A stiffness coeffcient is expressed in N/m and varies among episodes to equally cover the range $(300, 1400)$, which fits the real-world data range. A mass of all parts was adapted to the real values, as well as the mechanical impedance of objects, damping, and stiffness of all joints and springs in the system. Two IMUs are mounted on a MuJoCo's element called \textit{site} and located in the 3/4 of the length of each finger in the outside part of it. For experiments, we prepared two simulation datasets. The first one resembles the real-world data and consists of 5000 training-validation samples gathered from squeezing the box object only. We use it for an enrichment of real-world data. The second one was composed of objects in three different shapes - boxes, cylinders, and spheres. It counts 3999 training-validation samples -- 1333 samples per each object. In our research, it was used to verify whether the NN can avoid over-fitting to any particular shape. Additionally, to verify the NN performance among different shapes of objects we prepared 3 test datasets -- 133 samples for each object.

\subsection{Network Architecture}   

\begin{figure}[htbp]
\centerline{\includegraphics[width=0.45\textwidth]{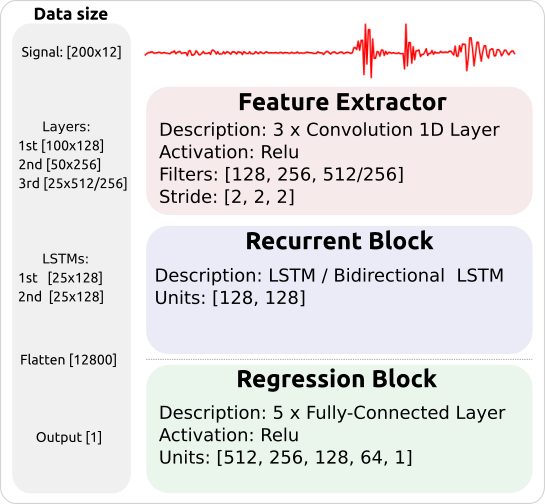}}
\caption{In our networks, the Feature Extractor produced high-level features from an input signal using 1D convolutions. In the ConvBiLstmNet and the ConvLstmNet, the Recurrent Block processed these features to find relevant connections for the stiffness estimation. However, in the former architecture it was done in the forward and backward manner (from the beginning of the signal and back). Finally, the Regression Block transformed high-level features into one scalar value. In our experiments, we exploited three architectures of neural networks. The difference is in the Recurrent Block -- both recurrent NNs have the reduced number of filters in the last convolutional layer and added LSTM cells with 256 units (2x128), while in the ConvNet the output from the Feature Extractor is passed directly to the Regression Block.} 
\label{fig:nn_scheme}
\end{figure}

Our neural networks predicted the stiffness parameter from fixed length sequences of accelerations and angular velocities measured by IMUs. In our research we proposed to test three types of neural networks -- the ConvNet based entirely on 1D convolutional blocks, the ConvLstmNet with forward LSTM units and the ConvBiLstmNet with bidirectional LSTM units. In both cases of LSTM-based NNs models, the recurrent part is placed after the convolutional block. In the end of each architecture we placed a fully-connected layer named the Regression Block. The scheme of the proposed neural networks architectures was depicted in the Figure~\ref{fig:nn_scheme}.

\textbf{Feature Extractor --} The neural network input was a standardized signal in the form of the 2-dimensional tensor. Each signal consisted of 12 time series with a length of 200. The main task of that block is to extract features while remaining in the time domain. Hence, data could be further processed recurrently or passed to the Regression Block directly. The Feature Extractor consisted of 3 consecutive 1D convolution layers with strides equal 2. In the ConvNet the number of filters was set to 128, 256, 512, while in the ConvLstmNet/ConvBiLstmNet, the last convolution block was reduced to 256 filters and replaced by the recurrent block with the same size.

\begin{figure}[htbp]
\centerline{\includegraphics[width=0.45\textwidth]{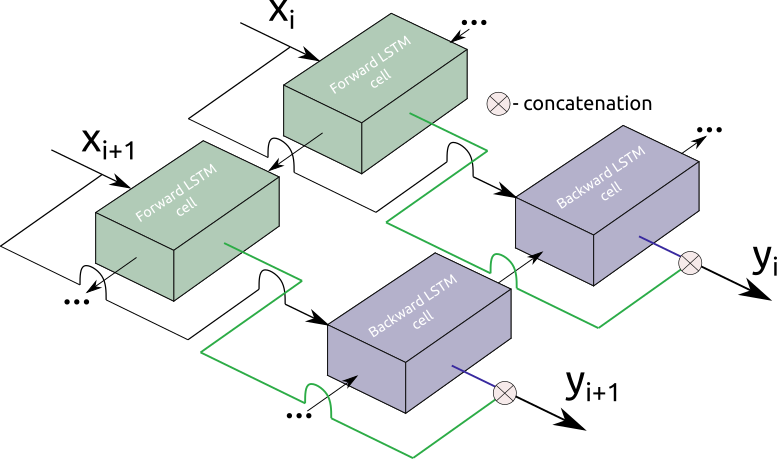}}
\caption{The core idea standing behind the bidirectional--LSTM used in the ConvBiLstmNet is as follows -- to prevent losing a context by the cell, process a sequence from the beginning to the end, do the same in the reversed direction and concatenate both \textit{passes}. Input $x_i$ refers to the $i$-th feature vector returned by the convolutional block.} 
\label{fig:rnn}
\end{figure}

\textbf{Recurrent Block --} It process high dimensional time series from the Feature Extractor in a recurrent manner using LSTM~\cite{LSTM}/bidirectional LSTM cells~\cite{BiDirRNN}. The input is mapped to a fixed-length vector that represents the entire signal in itself. In that way, we obtained a global, reduced description of the signal. Each recurrent cell consist of 128 units, as depicted in Figure \ref{fig:rnn}. In the the ConvLstmNet, both LSTM cells are organized in two sequential layers processing the signal in the forward direction only. Outputs of that block is finally forwarded to the Regression Block.

\textbf{Regression Block --} The last block was used to do a regression and output an estimated stiffness coefficient. The necessity of using a fully-connected block stems from the fact that extracted features and time dependencies between them are critical ingredients in the regression process, but they are not the answer itself. At the very end of the processing, it is necessary to transform the obtained features into stiffness coefficient estimate, which can be easily performed using the stack of fully-connected layers. The number of units in each layer remains unchanged for all tested architectures and is 512, 256, 128, 64, 1.

\section{Results}

The section of the results was divided as follows. Firstly, the simulation experiments were conducted. We verified which NN yielded the best performance on the simulation datasets and how well it was able to generalize among different shapes of squeezed objects. Secondly, the real-world experiments were presented using the NN architecture chosen during the simulation tests stage. We focused on the closing of the reality gap between simulation and real-world data. In all cross-validation experiments, we provided results obtained for the best epoch per each k-fold according to the MAPE. In all our experiments we used Adam optimizer with a learning rate set to 0.001. Each model was trained with the batch size 100 and all our solutions were trained for 100 epochs per each fold of the cross-validation.

\subsection{Neural Network Architecture Comparison}

First of all, in our experiments, we compared three types of neural networks and chose the best one for further experiments. The values of MAE/MAPE metrics from the cross-validation procedure were presented in Table~\ref{tab:1st}. The best performing network -- the ConvBiLstmNet, was chosen for further experiments.

\begin{table}[hbt]
\centering
\caption{The comparison of three NN architectures according to MAE/MAPE metrics. The usage of bidirectional LSTM units gave an improved performance comparing to the ConvNet and the ConvLstmNet .}
\begin{tabular}{c|c|c|c|c|c|c|}
\cline{2-7}
 & \multicolumn{2}{c|}{\textbf{ConvNet}} & \multicolumn{2}{c|}{\textbf{ConvLstmNet}} & \multicolumn{2}{c|}{\textbf{ConvBiLstmNet}} \\ \hline
\multicolumn{1}{|c|}{\textbf{k-fold}} & \textbf{MAE} & \textbf{MAPE} & \textbf{MAE} & \textbf{MAPE} & \textbf{MAE} & \textbf{MAPE} \\ \hline
\multicolumn{1}{|c|}{\textbf{I}} & 19,1 & 2,4 & 6,2 & 0,8 & 6,2 & 0,8 \\ \hline
\multicolumn{1}{|c|}{\textbf{II}} & 11,8 & 1,6 & 5,4 & 0,7 & 5,4 & 0,7 \\ \hline
\multicolumn{1}{|c|}{\textbf{III}} & 15,1 & 2,2 & 7,8 & 1,1 & 7,8 & 1,1 \\ \hline
\multicolumn{1}{|c|}{\textbf{IV}} & 14,6 & 1,9 & 6,7 & 0,9 & 6,7 & 0,9 \\ \hline
\multicolumn{1}{|c|}{\textbf{V}} & 18,1 & 2,1 & 6,2 & 1,0 & 6,2 & 1,0 \\ \hline
\multicolumn{1}{|c|}{\textbf{MEAN}} & \textbf{15,7} & \textbf{2,0} & \textbf{6,8} & \textbf{0,9} & \textbf{6,5} & \textbf{0,9} \\ \hline
\multicolumn{1}{|c|}{\textbf{STD DEV}} & \textbf{2,9} & \textbf{0,3} & \textbf{0,9} & \textbf{0,2} & \textbf{0,7} & \textbf{0,1} \\ \hline
\end{tabular}
\label{tab:1st}
\end{table}

\subsection{Shape Generalization}

To verify the capability of the ConvBiLstmNet to successfully estimate the stiffness parameter we conducted more experiments on the simulation-only datasets. We started the cross-validation procedure from scratch for chosen model and reported the MAE/MAPE for three different datasets in Table~\ref{tab:2nd}. Each test dataset was composed of signals from squeezing only one type of object so that the findings of the shape-dependent stiffness parameter regression could be provided.

\begin{table}[hbt]
\centering
\caption{The results from experiments on shape-invariant estimation of the stiffness parameter.}
\begin{tabular}{|c|c|c|c|c|c|c|}
\hline
\multirow{3}{*}{\textbf{k - fold}} & \multicolumn{6}{c|}{\textbf{Dataset}} \\ \cline{2-7} 
 & \multicolumn{2}{c|}{\textbf{Ball}} & \multicolumn{2}{c|}{\textbf{Box}} & \multicolumn{2}{c|}{\textbf{Cylinder}} \\ \cline{2-7} 
 & MAE & MAPE & MAE & MAPE & MAE & MAPE \\ \hline
\textbf{I} & 20,3 & 2,0 & 24,1 & 1,8 & 15,6 & 1,8 \\ \hline
\textbf{II} & 29,6 & 2,6 & 12,9 & 1,6 & 15,8 & 1,9 \\ \hline
\textbf{III} & 27,1 & 2,0 & 22,8 & 1,8 & 16,0 & 1,9 \\ \hline
\textbf{IV} & 21,8 & 2,1 & 17,7 & 16,6 & 18,4 & 1,9 \\ \hline
\textbf{V} & 19,3 & 2,0 & 24,4 & 1,5 & 20,8 & 1,9 \\ \hline
\textbf{MEAN} & \textbf{23,6} & \textbf{2,1} & \textbf{20,4} & \textbf{4,7} & \textbf{17,3} & \textbf{1,9} \\ \hline
\textbf{STD DEV} & \textbf{4,5} & \textbf{0,3} & \textbf{5,0} & \textbf{6,7} & \textbf{2,2} & \textbf{0,0} \\ \hline
\end{tabular}
\label{tab:2nd}
\end{table}

\subsection{Sim-To-Real Gap}

The central part of our research was about assessing the reality gap in the task of the stiffness parameter estimation. In that part of the experiments, we performed 5 training procedures of the ConvBiLstmNet on data with different number of real-life examples or noise added to simulation data, each composed of 5-fold cross-validation. In Table~\ref{tab:3rd} we reported MAE/MAPE metrics gathered while \textit{testing} each model on the separate dataset, not involved in the training/validation procedure. In the \textit{sim + noise} experiment we tried to close the reality gap, by adding a zero-mean Gaussian noise with standard deviation set to 0.7 for acceleration signals and 0.06 for the gyroscope readings. The parameters of the noise were adjusted by trials and errors, thus too large standard deviation resulted in the lack of the convergence ability of the NN, while too small caused model to over-fit to the simulation data and no clear rule for that phenomena is known. Each next cross-validation turn was performed on simulation datasets without noise and with a small number N of real-world data samples included in the training part. In Table~\ref{tab:3rd} we refer to them as \textit{sim + N real}.

\begin{table*}[hbt]
\centering
\caption{MAE/MAPE results reported for best epochs from each of the cross-validation turns. Introducing to the network even a small number of real-world signals resulted in a significant improvement in the performance.}
\begin{tabular}{|l|c|c|c|c|c|c|c|c|c|c|c|c|}
\hline
\multicolumn{1}{|c|}{\multirow{3}{*}{\textbf{\begin{tabular}[c]{@{}c@{}}Experiment\\ Name\end{tabular}}}} & \multicolumn{10}{c|}{\textbf{k - fold}} & \multicolumn{2}{c|}{\textbf{MEAN}} \\ \cline{2-13} 
\multicolumn{1}{|c|}{} & \multicolumn{2}{c|}{\textbf{I}} & \multicolumn{2}{c|}{\textbf{II}} & \multicolumn{2}{c|}{\textbf{III}} & \multicolumn{2}{c|}{\textbf{IV}} & \multicolumn{2}{c|}{\textbf{V}} & \multirow{2}{*}{\textbf{MAE}} & \multirow{2}{*}{\textbf{MAPE}} \\ \cline{2-11}
\multicolumn{1}{|c|}{} & MAE & MAPE & MAE & MAPE & MAE & MAPE & MAE & MAPE & MAE & MAPE &  &  \\ \hline
\textbf{sim + noise} & 281,3 & 37,7 & 275,0 & 38,5 & 275,6 & 38,4 & 282,7 & 37,6 & 256,6 & 37,9 & \textbf{274,2 $\pm$ 10,4} & \textbf{38,0 $\pm$ 0,4} \\ \hline
\textbf{sim + 50 real} & 190,6 & 23,1 & 216,1 & 27,1 & 187,8 & 26,4 & 151,8 & 21,6 & 200,7 & 27,7 & \textbf{189,4 $\pm$ 23,8} & \textbf{25,2 $\pm$ 2,7} \\ \hline
\textbf{sim + 100 real} & 134,6 & 20,6 & 108,3 & 17,6 & 134,9 & 19,6 & 126,8 & 18,6 & 126,6 & 18,3 & \textbf{126,2 $\pm$ 10,8} & \textbf{18,9 $\pm$ 1,2} \\ \hline
\textbf{sim + 150 real} & 89,3 & 12,9 & 85,9 & 13,7 & 92,7 & 13,2 & 73,9 & 11,0 & 79,9 & 10,2 & \textbf{84,3 $\pm$ 7,5} & \textbf{12,2 $\pm$ 1,5} \\ \hline
\textbf{sim + 200 real} & 66,9 & 9,1 & 49,3 & 7,0 & 82,6 & 10,9 & 67,4 & 8,4 & 56,6 & 8,0 & \textbf{64,6 $\pm$ 12,6} & \textbf{8,7 $\pm$ 1,5} \\ \hline
\end{tabular}
\label{tab:3rd}
\end{table*}

\section{Discussion}
\label{sec:discussion}

In the following section, we commented on the results and our observations for three types of experiments carried out in the course of our research.

\textbf{Architecture Choice --} We compared the performance of three types of neural networks in the task of a stiffness parameter estimation from IMUs readings, to choose the best one for the further analysis. All models were examined on the simulation dataset without real-world data samples. In Table~\ref{tab:1st} one can observe the results from cross-validation on the simulation dataset. The mean results of the MAE/MAPE show the advantage of the LSTM-based models in the performed task. The conclusions are twofold. Firstly, the ConvBiLstmNet is more accurate in its predictions than ConvNet, giving the result 6,5 N/m MAE and the 0,9\% MAPE, which means the improvement over 9,5 N/m and 1,1\% achieved by the ConvNet. Secondly, the stability of the learning process also improved and it can be observed in deviations of errors obtained between cross-validation folds. For ConvNet the standard deviation of results is 2,9 N/m MAE and 0,3\% MAPE, while the ConvBiLstmNet decreased these values to 0,9 N/m and 0,2\% respectively. Comparing two recurrent NNs, one can observe that the results are similar. However, the ConvBiLstmNet exhibits better performance in the MAE, what means than on average it made a lesser absolute error, hence that architecture was chosen for further experiments.

\textbf{Shape-Invariant Predictions --} To verify the generalization capability of the ConvBiLstmNet and verify its performance on different types of objects, we performed additional experiments. In Table~\ref{tab:2nd} we gathered the MAE/MAPE from testing the network on three separate datasets, each of which including only one type of object, while training on all shapes at once. All the results suggest that the proposed NN was able to generalize among different types of shapes and perform the shape-invariant stiffness parameter prediction. It appears that the cylinder-shaped objects are the easiest in the performed task, which is reflected in the lowest MAE/MAPE -- 17,3 N/m and 1,9\%. However, box objects gave the smaller values of MAE (20,4 N/m) than ball-shaped objects (23,6 N/m), while looking at the MAPE the situation was the opposite -- larger error was for boxes (4,7\% / 2,1\%). It means that the NN was inaccurate more often while estimating large stiffness values for boxes that resulted in the increased relative metric (MAPE), while for ball-shaped objects the quality of the estimation was decreased for small values that gave increased absolute measure (MAE). 

\begin{figure}[hbt]
    \centering
    \includegraphics[width=0.48\textwidth]{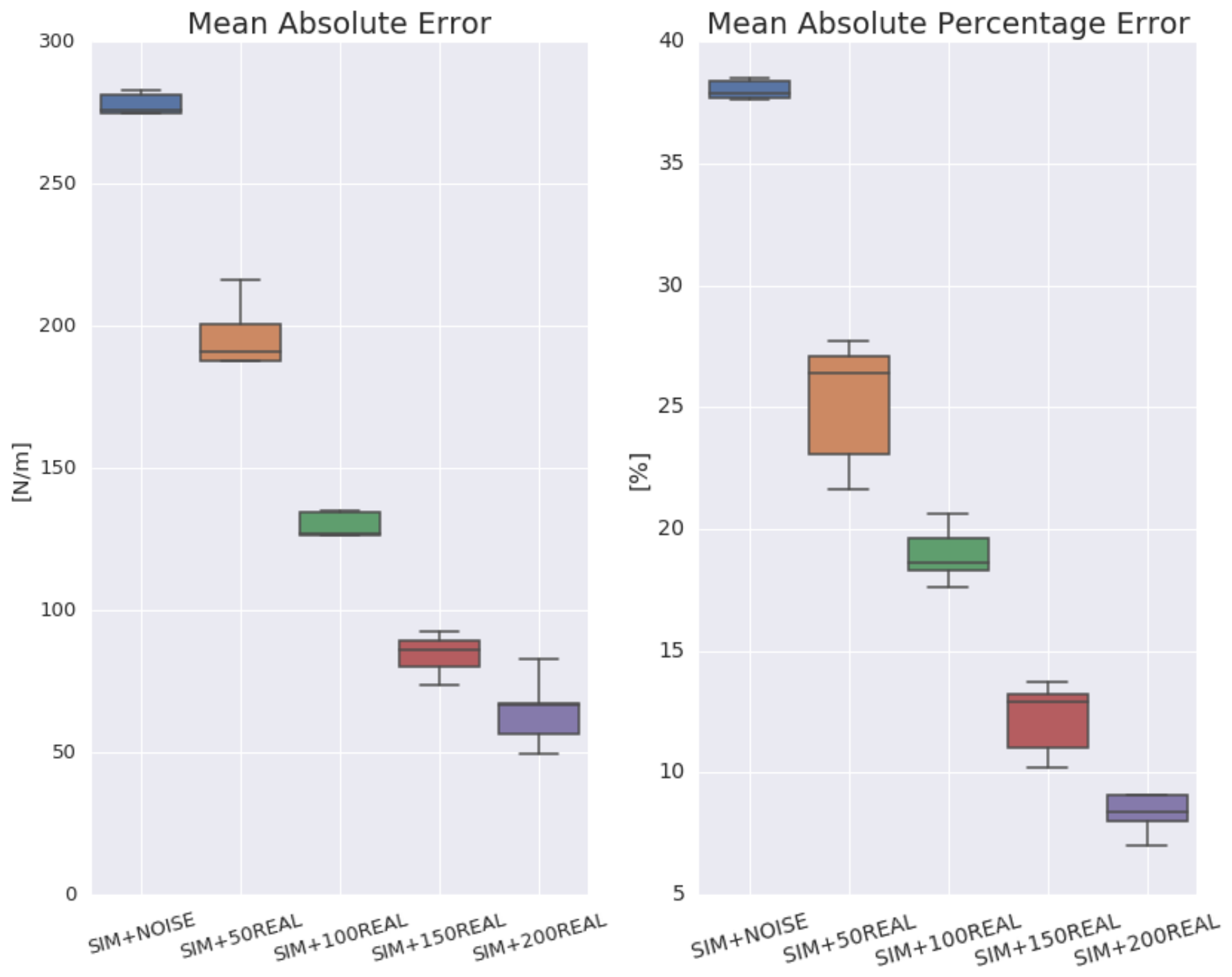}
    \caption{Results of MAE/MAPE from the testing on real-world data presented in the box plot. As the number of real data samples included in the training dataset increases, the test error decreases. Boxes represent consecutive experiments and consist of the five-number summary of the result (from the bottom of each box): minimum, first quartile, median, third quartile, and maximum value.}
    \label{tab:maemape}
\end{figure}

\textbf{Closing The Reality Gap --} In the task of haptic recognition of physical parameters, data from the physics simulator appeared to resemble the real-world signals only to some restricted extent. Although the results from \textit{sim + noise} tests were significantly worse than any of the \textit{sim + real} trail, the mean MAPE 38\% suggests that the correspondence between the simulation-only and real-world signals exists. Additionally, it is important to note that MAE/MAPE values from each fold in the \textit{sim + noise} experiment remained relatively close to each other, which means that the model prediction performance was similar for the entire dataset, as it was equally balanced in the stiffness parameters range. However, the reality gap cannot be considered as a solved problem, because the highest improvement was observed for experiments with the real-world signals included in the training dataset. In Figure~\ref{tab:maemape} one can observe the decreasing value of MAE/MAPE metrics as the number of real data samples are added to the training dataset. In our experiments we do not include the results from the training on the real-world data only, as they would be incomparable with other experiments, due to the low variability of the stiffness coefficient. Additionally, the number of data samples would be too low to assess the fair comparison in the real-world scenario.  The lowest MAE/MAPE obtained in experiments on closing the reality gap was achieved for \textit{sim + 200 real} trial and was 64,6 N/m and 8,7\%. However, in the \textit{sim + 50 real} experiment, the added number of real samples constituted only 1,2\% of the entire training dataset, but the largest performance improvement among all experiments was observed. The improvement was 84,8 N/m and 12,8\% of the MAE/MAPE.

\section{Conclusions}

We have shown that estimation of the object's physical parameters using data from IMU sensors is possible and beneficial due to the low cost of setup and no further need for sophisticated equipment. Our deep learning solution solves a problem of the stiffness estimation in the soft robotics area, introducing a novel approach, which associates an embodied and artificial intelligence. Their combination may lead to a system robust to unforeseen and changing external conditions. While currently used methods of stiffness search exploit techniques of measurement or direct estimation, the method proposed by us is characterized by the discovery of knowledge and causal relationships related to the characteristics of a given object and its physical features. Research on the discovery of knowledge acquired by neural networks may result in the diagnosis of the intuition behind the natural behavior of humans in the tasks of manipulating objects. We find it likely that similar solutions, based on low-cost sensors and deep learning, may be successfully applied for robotic manipulation in everyday scenarios. We hope that the published data and the implementation of neural networks used in our experiments will inspire other researchers to delve into the area of soft grippers and perception of the physical world based on tactile data in robotics.

\section*{Acknowledgment}
This work is supported by grant No. LIDER/3/0183/L-7/15/NCBR/2016 funded by The National Centre for Research and Development (Poland).

\bibliographystyle{plain}
\bibliography{root}

\end{document}